\documentclass[letterpaper]{article} 
\usepackage{aaai25}  
\usepackage{times}  
\usepackage{helvet}  
\usepackage{courier}  
\usepackage[hyphens]{url}  
\usepackage{graphicx} 
\usepackage{natbib}  
\usepackage{caption} 
\usepackage{booktabs}
\usepackage[most]{tcolorbox} 
\usepackage{subcaption}
\usepackage{algorithm}
\usepackage{algorithmic}

\usepackage{newfloat}
\usepackage{listings}
\DeclareCaptionStyle{ruled}{labelfont=normalfont,labelsep=colon,strut=off} 
\floatstyle{ruled}
\newfloat{listing}{tb}{lst}{}
\floatname{listing}{Listing}
\title{APT: Architectural Planning and Text-to-Blueprint Construction Using Large
Language Models for Open-World Agents}
\author{
    Jun Yu Chen,
    Tao Gao
}
\affiliations{
    University of California, Los Angeles\\
    Los Angeles, CA, USA\\

}

\usepackage{bibentry}

\begin{document}

\maketitle

\begin{abstract}
We present APT, an advanced Large Language Model (LLM)-driven framework that enables autonomous agents to construct complex and creative structures within the Minecraft environment. Unlike previous approaches that primarily concentrate on skill-based open-world tasks or rely on image-based diffusion models for generating voxel-based structures, our method leverages the intrinsic spatial reasoning capabilities of LLMs. By employing chain-of-thought decomposition along with multimodal inputs (textual and visual), the framework generates detailed architectural layouts and blueprints that the agent can execute under zero-shot or few-shot learning scenarios. Our agent incorporates both memory and reflection modules to facilitate lifelong learning, adaptive refinement, and error correction throughout the building process. To rigorously evaluate the agent’s performance in this emerging research area, we introduce a comprehensive benchmark consisting of diverse construction tasks designed to test creativity, spatial reasoning, adherence to in-game rules, and the effective integration of multimodal instructions. Experimental results using various GPT-based LLM backends and agent configurations demonstrate the agent’s capacity to accurately interpret extensive instructions involving numerous items, their positions, and orientations. The agent successfully produces complex structures complete with internal functionalities such as Redstone-powered systems. A/B testing indicates that the inclusion of a memory module leads to a significant increase in performance, emphasizing its role in enabling continuous learning and the reuse of accumulated experience. Additionally, the agent's unexpected emergence of scaffolding behavior highlights the potential of future LLM-driven agents to utilize subroutine planning and leverage emergence ability of LLMs to autonomously develop human-like problem-solving techniques.
\end{abstract}

\begin{links}
    \link{Code}{https://github.com/spearsheep/APT-Architectural-Planning-LLM-Agent}
\end{links}

\section{Introduction}
In recent years, autonomous agents within the Minecraft environment have become a focal point of research, with methods like reinforcement learning\cite{baker2022video} and large language models (LLMs) playing a central role\cite{wang2024voyager,fan2022minedojo,yuan2023skillreinforcementlearningplanning}. These methods allow agents to learn by engaging directly with the game world, making real-time decisions, and adapting based on accumulated experiences. The aim is to enable these agents to develop lifelong learning capabilities, improving their skills continuously and handling increasingly complex tasks over time. Yet, while existing research demonstrates potential, it often neglects nuanced creativity and spatial reasoning needed for more sophisticated construction tasks that reflect human-like abilities.

Current research predominantly concentrates on single agents performing straightforward tasks aligned with Minecraft’s technology tree, such as tool crafting, mining, and item productions\cite{wang2023jarvis,yu2024minelandsimulatinglargescalemultiagent}. While these tasks are essential to game mechanics, they follow predictable, task-driven pathways and do not inherently challenge the agent to imagine or create complex building structures. For instance, building houses or farms requires precise block placement in a three-dimensional space with spatial reasoning, which is a far more intricate process than crafting tools or mining resources. Constructing these complex structures involves long-term planning, the ability to envision an architectural blueprint, and a sequential building execution that current agent systems typically lack.

Moreover, while a few projects have integrated LLMs to enhance agent creativity, their generative capacities are limited to the pre-existing Minecraft knowledge embedded in online data. LLMs can replicate known structures but struggle with original designs or untrained configurations, limiting their adaptability in constructing unique or complex architectures, such as villages or multi-room buildings. Thus, despite their generative strengths, LLMs are often unable to build beyond the constraints of their training data and adapt to unfamiliar domains\cite{Ahn2022DoAI}, which hinders them from achieving the originality and spatial reasoning required for elaborate construction projects.

\begin{figure*}[ht]
    \centering
    \includegraphics[width=\textwidth]{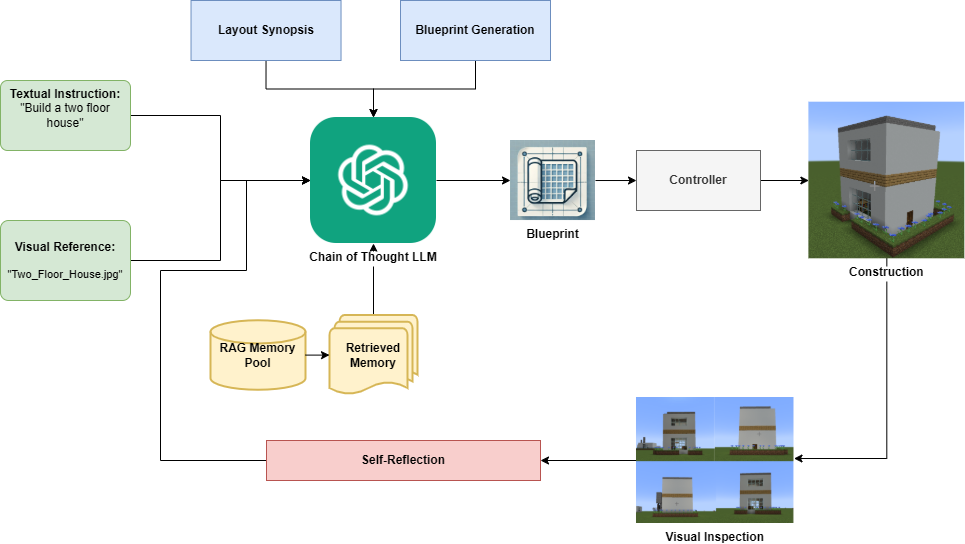} %
    \caption{Agent workflow for open-ended construction tasks. The agent begins with either textual instructions, a visual reference, or both. These inputs are processed through a Chain-of-Thought (CoT) module, which first generates a layout synopsis and subsequently produces the blueprint code.Relevant construction memories retrieved from the RAG memory pool are also integrated into the CoT module to enhance planning. The blueprint is then executed by the controller, which utilizes primitive actions to perform the construction within the environment. In cases of unsuccessful executions, the agent employs visual inspection and self-reflection to identify and correct errors, iteratively refining its construction plans in a closed-loop process.}
    \label{fig:agent_workflow} 
\end{figure*}

\subsection{Related Work on Structure-Building Agents}

\subsubsection{Diffusion Model for Creation}
Although research on structure-building agents is relatively unexplored, one notable study demonstrates an innovative approach by utilizing embodied agents that leverage a diffusion model as an "imagination" to generate voxel-based images of structures\cite{zhang2023creativeagentsempoweringagents,9009507}. A trained controller then translates these visual outputs into construction actions. This method highlights the potential of integrating generative models in autonomous building tasks. However, it relies heavily on image-based diffusion models\cite{Rombach2021HighResolutionIS}, where the construction process begins with generating a visual representation of the structure. This dependency, while effective for capturing aesthetic designs, leaves room for exploring alternative approaches.

In particular, leveraging the innate capabilities of Large Language Models (LLMs) to directly map textual instructions to structured blueprints offers a promising direction. Such a direct mapping bypasses the intermediate step of visual generation, potentially streamlining the process and enhancing efficiency in design creation. By utilizing LLMs for zero-shot or few-shot blueprint generation, agents could simplify complex workflows and expand their adaptability across diverse building tasks\cite{wei2022finetuned}.

\subsubsection{Interior Design and Spatial Representation}
The reliance on a single exterior image for voxel-based blueprint generation introduces challenges in accurately representing internal spatial configurations and item placements. For functional structures requiring precise layouts—such as redstone wiring or detailed item orientations—capturing logical spatial relationships becomes critical. While diffusion models excel in generating stylistic and creative exterior designs, they often struggle to incorporate the fine-grained interior details necessary for replicating realistic architectural functionality. This opens an avenue for approaches that integrate spatial reasoning and logical representation to address both the exterior and interior requirements of complex structures.
\subsubsection{Lack of Standardized Benchmarks
}
As the field of autonomous structure-building agents is emerging, a lack of standardized benchmarks poses a challenge for objective performance evaluation. Developing and implementing standardized benchmarks would allow for more consistent skill assessment and comparative analysis across different architectures and methodologies.

\subsection{Research Objectives}
We propose APT, a few-shot learning framework designed to develop LLM-based agents capable of constructing intricate structures in Minecraft by interpreting both textual instructions and visual references. APT emphasizes creativity and accuracy in structure building, aiming to push the boundaries of what AI-driven agents can achieve in this domain. The key objectives of our study are as follows:

\begin{enumerate}
    \item \textbf{Exploring LLM Capabilities for Structure Building:}  
    We aim to leverage advances in recent GPT models to assess their ability for strategic and creative envisioning. Specifically, we will explore how LLMs can translate complex textual instructions directly into blueprint-level details, enabling precise and innovative structure designs.
    \item \textbf{Incorporating Memory and Reflection Mechanisms:}  
    To foster adaptive and lifelong learning, we plan to implement memory and reflection modules. These mechanisms will allow agents to learn from past building experiences, refine their strategies, and improve performance over time, contributing to more dynamic and intelligent behavior[9,20].
    \item \textbf{Establishing Comprehensive Benchmarks:}  
    To address the lack of standardized evaluation metrics in this domain, we propose a robust benchmark for structure-building tasks. This benchmark will rigorously evaluate agents across multiple skill areas, including instruction interpretation, creativity, adherence to game rules, and integration of visual inputs.
\end{enumerate}

Through this research, we aim not only to advance the capabilities of autonomous agents in Minecraft but also to contribute to the broader field of creative, task-oriented agent performance assessment by establishing more robust evaluation standards.

\begin{figure*}[!ht]
    \centering
    \includegraphics[width=0.8\textwidth]{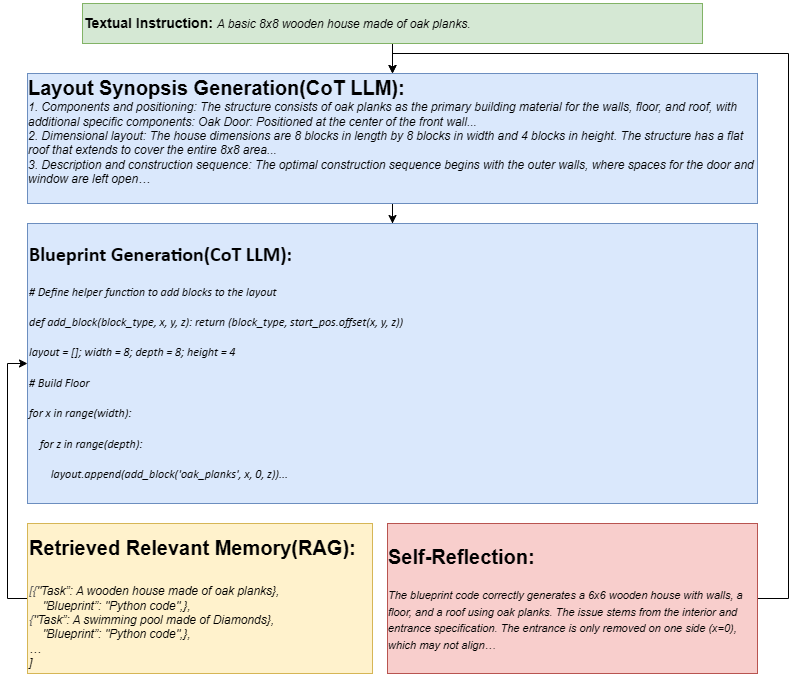} %
    \caption{Agent Workflow Example: Crafting a Simple Wooden House Task.}
    \label{fig:workflow_example} 
\end{figure*}

\section{APT Multi-modal LLM Agent Architecture}
In this section, we introduce the architecture of the APT agent, as illustrated in Figure~\ref{fig:agent_workflow}, emphasizing its modular design and workflow. We demonstrate how LLMs leverage spatial reasoning and planning to generate structure blueprints from textual instructions, with enhancements provided by memory and reflection mechanisms. Additionally, Figure~\ref{fig:workflow_example} showcases example outputs from each component of the workflow, offering a detailed perspective on the agent’s operation.
\subsection{Multimodal Instruction Processing}
Our agent leverages the GPT-4o LLM backend to interpret and respond to both textual and visual instructions, allowing for versatile task inputs‌\cite{gpt4o,gpt4_technical_report}. For example, in a task such as building a wooden house, the agent can receive a detailed textual description specifying components and dimensions, a reference image of the house, or both. This multimodal capability enhances the agent’s adaptability and accuracy, as it can process comprehensive information in different formats to initiate the building task in Minecraft.
\subsection{Chain-of-Thought(CoT) Module}
The Chain-of-Thought (CoT) module is central to our agent's architecture, designed to decompose complex tasks into manageable steps for improved reasoning and accuracy \cite{chen2023chatcot}. As shown in Figure~\ref{fig:agent_workflow}, the module consists of two primary components: Architecture Layout Synopsis and Blueprint Generation.
\subsubsection{Architecture Layout Synopsis}
We apply prompt engineering in this step to guide the LLM in breaking down the input (image or text) into structured information: components and positioning, dimensional layout, description, and construction sequence, as shown in Figure~\ref{fig:workflow_example}. The components and positioning output identifies items within the structure and their spatial relations, focusing particularly on interior designs. The dimensional layout provides an overall size and shape specification for the structure. The description and construction sequence outlines the structure's purpose and design while detailing a logical order for building—such as completing floors before upper levels and roofs before interior furnishings. This structured approach reduces the chances of execution errors, such as inaccessible areas, and optimizes building efficiency by prioritizing foundational elements first.
\subsubsection{Blueprint Generation}
Following the architecture layout synthesis, the structured information feeds into the second phase of the Chain-of-Thought (CoT) module: blueprint generation. In this step, the agent is prompted to generate Python code that generates a 3D layout blueprint, as depicted in~\ref{fig:workflow_example}. The blueprint is represented as a list of tuples, where each tuple specifies the block type and its exact 3D coordinates within the Minecraft environment. This approach directly engages the spatial reasoning and planning capabilities of the LLM, allowing it to map high-level instructions into detailed spatial and positional representations. By working in a structured text-to-blueprint paradigm, the agent bypasses the intermediate image-generation step, reducing complexity, training requirements, and potential sources of error.

\subsubsection{Construction via Primitive Actions}
The final execution of the blueprint occurs through a sequence of primitive actions facilitated by the Mineflayer library\cite{mineflayer2024}, including pathfinding and block manipulation functions adapted from Voyager's primitive action list\cite{wang2024voyager}. Our approach emphasizes fundamental actions—such as placeBlock, jump, pathfinding, and mineBlock—to construct the structure directly. This contrasts with other agents, such as Jarvis-1, which use MineDojo controllers for higher-level construction management\cite{fan2022minedojo}. 

\subsubsection{Retrieval-Augmented Planning}
To enable learning from past tasks, we incorporated a memory pool supported by a Retrieval-Augmented Generation (RAG) system‌\cite{NEURIPS2020_1457c0d6}, using Chroma vector databases in LangChain. This memory architecture allows the agent to retrieve similar past plans when working on new tasks, avoiding repeated errors and enhancing task efficiency. Retrieval queries use cosine similarity search, with the top-k results retrieved as contextual input for blueprint generation, allowing the agent to adapt prior successful strategies to current challenges. As illustrated in Figure~\ref{fig:workflow_example}, multiple past construction memories are stored as key value pairs, with the most relevant one— in this case, a wooden house made of oak planks—being retrieved.
\subsubsection{Self-Reflection and Error Correction}
As depicted in Figure~\ref{fig:agent_workflow} and~\ref{fig:workflow_example}, our agent includes an optional self-reflection loop, designed to address execution errors or suboptimal constructions\cite{shinn2023reflexionlanguageagentsverbal}. In case of errors, the agent can perform a visual inspection of the structure through multiple first-person screenshots taken from various perspectives, facilitated by Prismarine viewers\cite{mineflayer2024}. These images, combined with the original instructions and the execution code, are reprocessed through the CoT module to refine the blueprint and identify specific issues‌\cite{wei2022chain}. The agent's reflective loop parameter can be adjusted to control the degree of error-checking.

\section{Benchmark Tasks}

To rigorously evaluate our agent's construction capabilities in Minecraft, we have developed a benchmark consisting of five structured construction tasks. Each task is crafted to test specific skills, such as spatial reasoning, adherence to game rules, creativity, and accuracy in interpreting instructions\cite{10610855}. These tasks range in complexity, from simple builds to intricate designs with functional components, providing a comprehensive assessment of the agent’s architectural and creative abilities.

\begin{enumerate}
    \item \textbf{Simple Wooden House}
    \begin{itemize}
        \item \textbf{Description:} A basic 8x8 wooden house equipped with a door, window, bed, and crafting table inside.
        \item \textbf{Skills Tested:} This task assesses the agent's ability to execute fundamental architectural planning and manage basic interior furnishing by correctly orienting and positioning multiple items within a confined space.
    \end{itemize}

    \item \textbf{Snow Pyramid}
    \begin{itemize}
        \item \textbf{Description:} A pyramid structure made entirely of snow and ice blocks.
        \item \textbf{Skills Tested:} This task primarily challenges the agent's creativity by requiring the use of unconventional building materials. Unlike typical pyramid builds using sand, the agent must design a pyramid with snow and ice—materials not traditionally associated with pyramids—without relying on prior construction knowledge or standard block associations.
    \end{itemize}

    \item \textbf{Village House (from Reference Image)}
    \begin{itemize}
        \item \textbf{Description:} A small but structurally complex village house based on a reference image.
        \item \textbf{Skills Tested:} This benchmark evaluates the agent’s ability to interpret and replicate a design from a visual reference. The agent must pay attention to fine details, ensuring the final structure closely matches the reference. This task is crucial for assessing the agent's skill in translating visual information into a precise 3D construction, an essential ability for tasks requiring detailed replication of human-created designs.
    \end{itemize}

    \item \textbf{Watchtower with Redstone Lighting System}
    \begin{itemize}
        \item \textbf{Description:} A tall watchtower featuring a Redstone-powered lighting system at the top, designed to illuminate automatically at night.
        \item \textbf{Skills Tested:} This task tests the agent’s ability to manage vertical space, implement functional Redstone mechanics, and interact dynamically with the environment (day-night cycle). Success here requires a solid understanding of wiring mechanisms and game rules; agents lacking this knowledge will struggle to build a functional lighting system.
    \end{itemize}

    \item \textbf{Two-Floor Mansion}
    \begin{itemize}
        \item \textbf{Description:} A two-story mansion with intricate interior layouts, including room plans, a flower garden, and a chimney.
        \item \textbf{Skills Tested:} This advanced task challenges the agent to handle a complex design with multiple floors and detailed interior layouts, demanding precise block placement and long-horizon planning abilities. The agent must interpret a visual blueprint to achieve a cohesive design that maintains functionality and aesthetic appeal across multiple spaces.
    \end{itemize}
\end{enumerate}

\begin{figure}[h]
\centering
\includegraphics[width=0.45\textwidth]{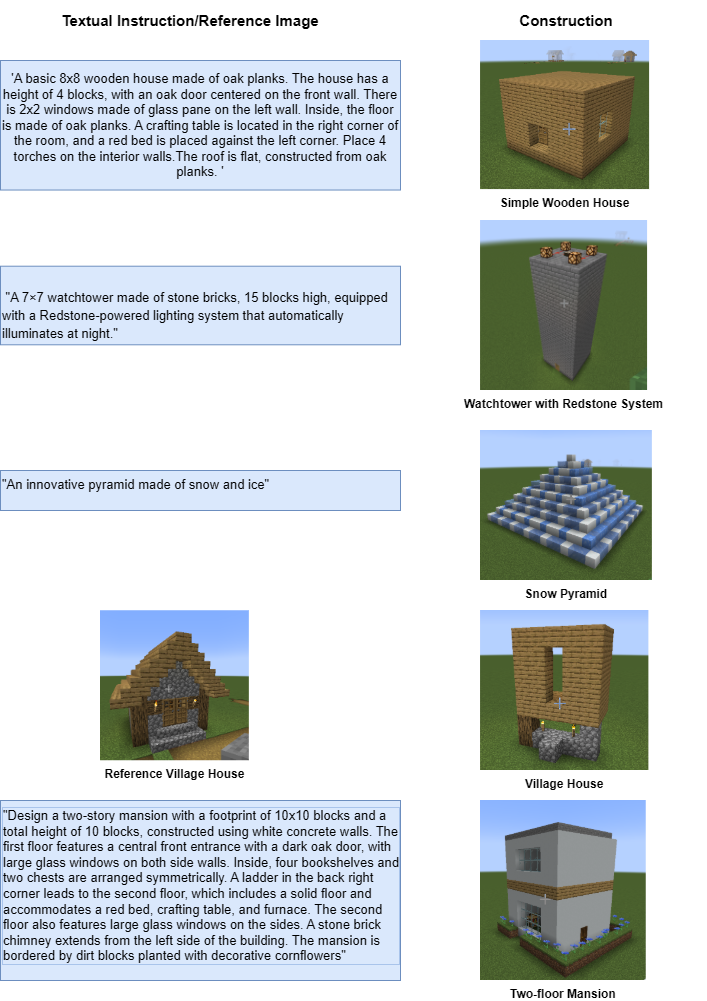}
\caption{Benchmark set structures constructed by our APT agent based on provided textual instructions or visual references. Some descriptions are highly detailed and lengthy, specifying precise item placements to rigorously evaluate the LLM's ability to imagine and reason spatially.}
\label{fig:benchmark_image}
\end{figure}

\begin{figure}[hb]
\centering
\includegraphics[width=0.45\textwidth]{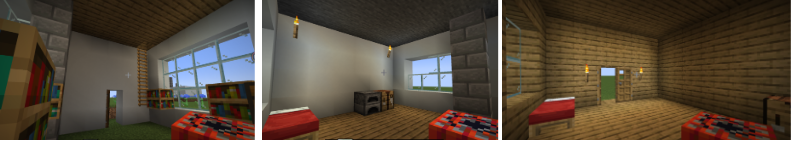}
\caption{Interior views, from left to right: the first floor of the Two-floor Mansion, the second floor of the same mansion, and the interior room of the Simple Wooden House.}
\label{fig:interior_structure}
\end{figure}

\begin{figure*}[ht]
    \centering
    \begin{subfigure}[t]{0.48\textwidth} 
        \centering
        \includegraphics[width=\textwidth]{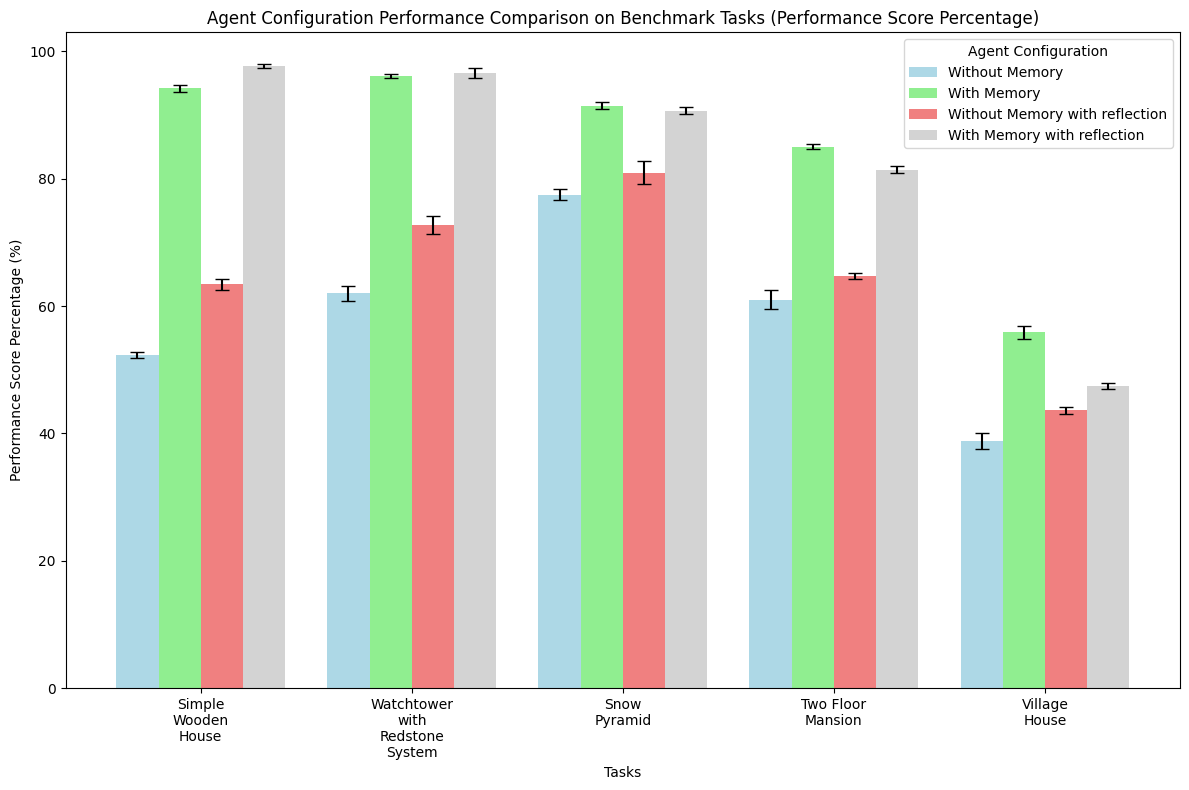}
        \caption{Comparison of performance score percentages of APT agent on benchmark tasks using different configurations, with variations in the use of memory and reflection modules.}
        \label{fig:agent_configuration}
    \end{subfigure}
    \hfill
    \begin{subfigure}[t]{0.48\textwidth} 
        \centering
        \includegraphics[width=\textwidth]{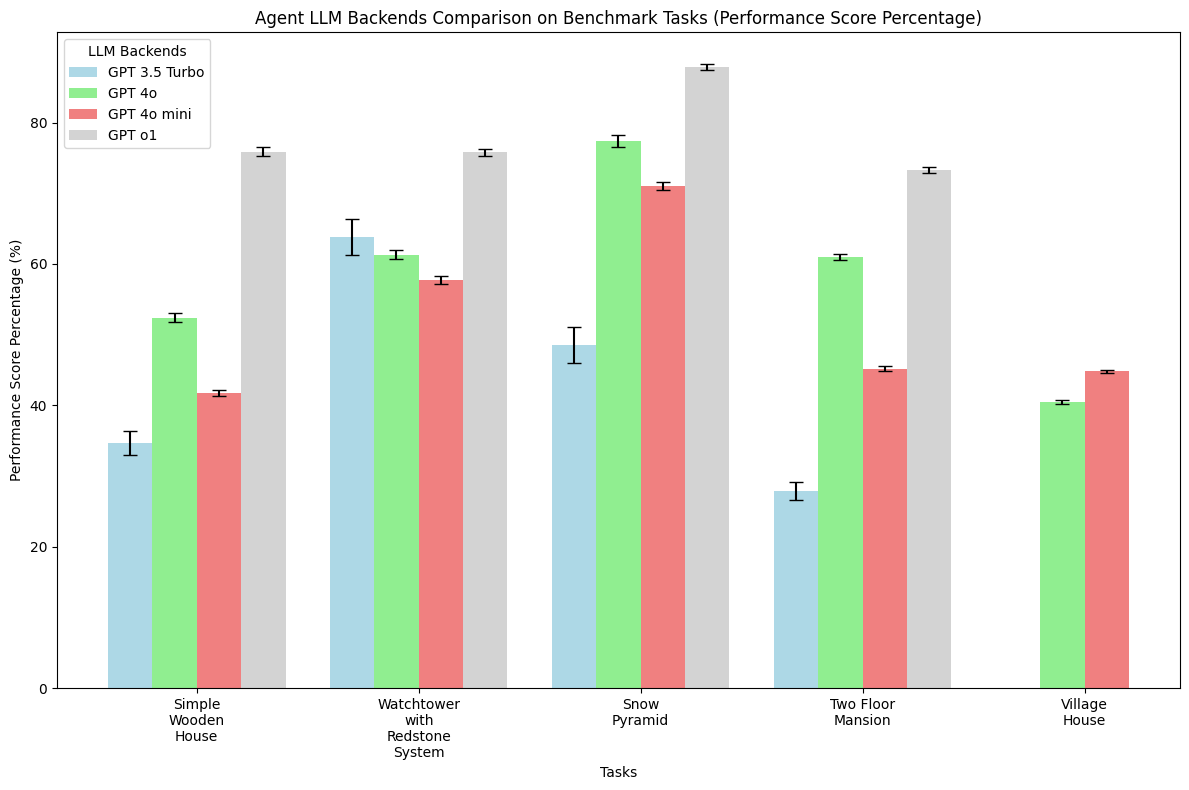}
        \caption{Comparison of performance score percentages of APT agent on benchmark tasks using differentLLM backends.}
        \label{fig:agent_backend}
    \end{subfigure}
    \caption{Performance score percentages based on agent configurations (left) and LLM backends (right).}
    \label{fig:bar_graphs}
\end{figure*}
\subsection{Evaluation Framework (VLM Evaluation)}
Quantitatively evaluating the aesthetics and accuracy of architectural constructions in Minecraft is a significant challenge, as item placement often requires dynamic weighting, leading to potential inconsistencies and a lack of objective measures for creativity and aesthetics. Traditional evaluation methods rely on human assessment, which, while thorough, is labor-intensive and can be influenced by subjective preferences‌\cite{baker2022video}. To address this, we leverage advanced Vision-Language Models (VLMs) such as GPT-4o, renowned for their capabilities in vision-language reasoning and interpretation, to automate the evaluation of our agent’s constructed architectures\cite{zhang2023creativeagentsempoweringagents}. A sample evaluation template, showcasing these criteria, is provided below:

\begin{tcolorbox}[
    colback=gray!5!white,     
    colframe=black!75!white, 
    title=Prompt for Evaluation, 
    sharp corners,           
    fonttitle=\bfseries,breakable      
]
\label{eval_prompt} 
{Your task is to evaluate the building across four key aspects:}
\begin{enumerate}
    \item \textbf{Correctness:} How accurately does the building adhere to the provided instructions, accounting for the inclusion of all specified components, block placements, and overall structure shape?
    \item \textbf{Complexity:} How intricate and detailed is the structure?
    \item \textbf{Creativity:} How unique and imaginative is the design?
    \item \textbf{Functionality}: How well does the building serve its intended purpose or function?
\end{enumerate}
{Please provide a score (out of 10) for each of these aspects. Additionally, provide an overall total score based on the individual aspect ratings.}

\textbf{Instruction:} \texttt{\{INSTRUCTION\}} \\
\textbf{Image of the building:} \texttt{\{IMAGE\}}
\end{tcolorbox}

\subsection{Evaluation Metrics}
As shown in the prompt for evaluation, our evaluation framework is based on four primary metrics: Correctness, Creativity, Complexity, and Functionality. Not all metrics apply uniformly to every task, as each construction task is designed to test specific skill sets. For example, the "Simple Wooden House" task emphasizes the agent’s ability to accurately follow instructions, so it is primarily assessed on Correctness. Evaluating it on Complexity would be unfair, as the structure itself is designed to be straightforward and simple.

\section{Results and Analysis}
With the integration of memory and self-reflection modules, the APT agent with GPT-4o backend achieved notable performance scores across most tasks: 97.8\% on the Simple Wooden House, 96.7\%  on the Watchtower with Redstone Lighting System, 90.7\%  on the Snow Pyramid, and 81.4\% on the Two-Floor Mansion. As depicted in Figure~\ref{fig:benchmark_image} and Figure~\ref{fig:interior_structure}, the agent's creations for these four tasks demonstrate a commendable level of adherence to the provided instructions. However, performance on the Village House with Reference Image task was significantly lower, with an average score of 47.5\%. As depicted in Figure~\ref{fig:agent_configuration}, adding memory and reflection modules yielded only minimal improvements for this task.

As shown in Figure~\ref{fig:benchmark_image}, the construction of "Village House" by our APT agent replicates only the individual items depicted in the reference, including the use of torches, cobblestone, oak wood, and oak planks. However, it fails to capture the three-dimensional structure of the reference image. This task required the agent to interpret a reference image accurately, suggesting that visual fidelity and detailed reproduction remain challenging for LLM-based agents, especially in zero-shot or few-shot scenarios. This indicates that reasoning from images to descriptions and then to blueprints is still a complex hurdle for LLMs, regardless of the configuration used.

\subsection{Ablation Experiment on Agent with Memory and Self-reflection}
We conducted ablation experiments to systematically compare various agent configurations (with and without memory and reflection) using a GPT-4o backend, assessing the individual and combined contributions of each module to overall performance in Figure~\ref{fig:agent_configuration}. The results demonstrate that the memory module significantly enhances the APT agent’s performance, with an average increase of 47.3\% in performance scores when memory is enabled. Even with the addition of the reflection module, the APT agent’s performance still saw a substantial boost, with a combined increase of 26.24\%. In contrast, without memory, the implementation of reflection alone improved performance by 12.8\%, suggesting that while reflection yields moderate gains\cite{huang2024large}, memory remains the more influential factor in enhancing agent capabilities.

To evaluate robustness, we calculated the standard deviation of overall scores across all test tasks, each assessed over 10 trials. Standard deviation here reflects consistency, with larger values indicating less robust performance. Notably, it is evident from Figure~\ref{fig:agent_configuration} that configurations lacking memory (either with or without reflection) exhibited higher standard deviation, underscoring the variability of LLM-generated plans and reasoning paths even with the temperature set to zero, which was intended to produce more predictable outputs. In contrast, implementing RAG with memory reduced variability, enabling the model to generate more stable and deterministic plans.

\subsection{Agent Workflow with Different LLM Backend}
As depicted in Figure~\ref{fig:agent_backend}, We evaluated the APT agent's performance across five construction tasks using various LLM backends: GPT-3.5 Turbo, GPT-4o, GPT-4o Mini, and GPT-o1. Each backend demonstrated unique capabilities, with significant performance differences across tasks. Notably, GPT-o1 achieved the highest overall average score of 78.22\%, compared to GPT-3.5 Turbo at 43.75\%, GPT-4o at 63.02\%, and GPT-4o Mini at 53.94\%. This outcome suggests that GPT-o1 exhibits superior reasoning abilities, particularly in handling complex spatial structures and accurately interpreting detailed textual instructions.

In Figure~\ref{fig:agent_backend}, we observed that GPT-3.5 Turbo struggled more with tasks such as the "Simple Wooden House" and "Two-Floor Mansion," both of which required understanding intricate spatial configurations and processing extended textual instructions. Moreover, because GPT-3.5 Turbo and GPT-o1 lack multimodal capabilities, the “Village House” task was only tested on GPT-4o and GPT-4o Mini. Both models performed poorly on this visually-driven task, underscoring current limitations in LLMs for tasks that demand high visual fidelity and nuanced spatial reasoning.

\begin{figure}[h]
    \raggedright
    \includegraphics[width=0.45\textwidth]{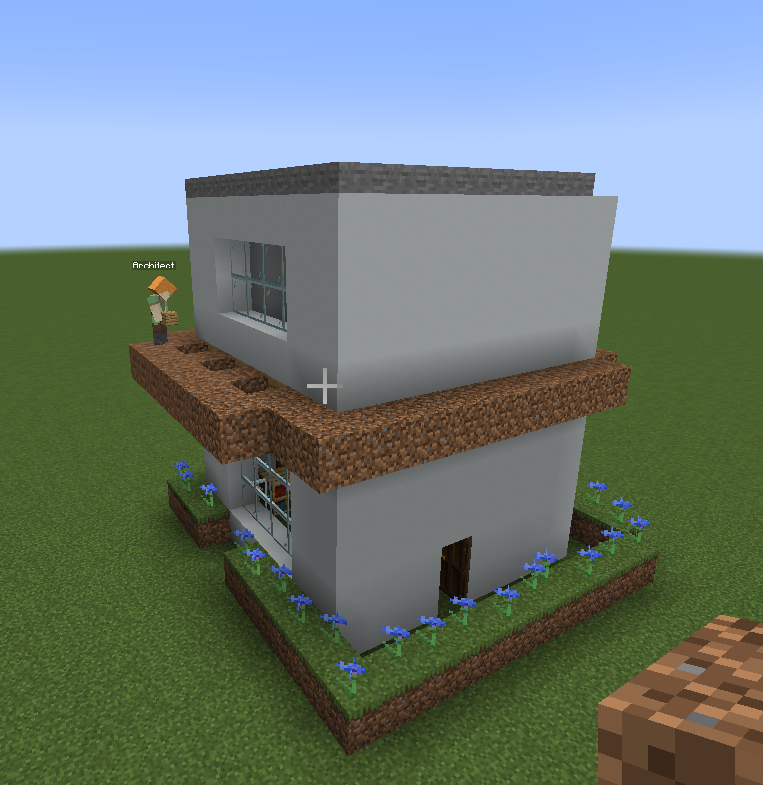} 
    \caption{Our APT agent demonstrates emergent abilities in constructing scaffolding as a part of the blueprint execution plan.}
    \label{fig:scaffolding}
\end{figure}

\subsection{Scaffolding ability}
One unexpected finding in agent behavior was that some blueprint layout code generated by our chain-of-thought modules included scaffolding as part of the construction sequence, as shown in Figure~\ref{fig:scaffolding}. In Minecraft, where agents are restricted from flying, scaffolding (e.g., stacking dirt or temporary blocks) is commonly used by players to reach higher floors and build efficiently. As observed in This behavior in our APT agent emerged without any explicit prompt for scaffolding, suggesting that the LLM inferred this technique while "imagining" an optimal construction sequence. This phenomenon is intriguing as it highlights the LLM's potential capability for implicit reasoning and common sense knowledge—recognizing and incorporating practical construction strategies from inferred context rather than direct instructions\cite{zhao2023large}.

\begin{table}[ht]
\centering
\begin{tabular}{@{}lr@{}}
\toprule
Correlation Type           & Coefficient \\ \midrule
Pearson                    & 0.988      \\
Spearman                   & 0.903      \\ \bottomrule
\end{tabular}
\caption{Correlation coefficients between Human and Machine evaluation. Pearson's coefficient measures strength of linear correlation, while Spearman's coefficient measures monotonic correlation based on ranks.}
\label{tab:correlation_metrics}
\end{table}

\subsection{Consistency between Human Evaluation and AI Evaluation}
To ensure that the evaluation of the VLM aligns closely with human judgment, we invited 22 participants to manually assess the structures created by our agents. Recognizing the critical influence of the memory component on the agent's performance, we focused the evaluation on structures generated by agents using the GPT-4o backend, comparing outputs both with and without memory. The reflection component was excluded from this assessment, as the memory component plays a more substantial role in shaping the overall performance of the agent.

To evaluate the consistency and alignment between the VLM evaluation and human judgment, we utilized the Pearson correlation coefficient and Spearman correlation coefficient, which are standard metrics for assessing the strength and nature of relationships between two sets of data \cite{schober2018correlation}. The results in Table~\ref{tab:correlation_metrics} show a Pearson correlation coefficient of 0.988 and a Spearman correlation coefficient of 0.903, indicating strong alignment and reliability between human assessments and the VLM evaluation. These findings validate the VLM evaluation process as a dependable alternative to human judgment for assessing agent-generated structures.

\subsection{Limitations and Future Directions}
One limitation of our study is the constrained memory capacity of the agent. Expanding the memory pool with a broader and more diverse set of building experiences could improve retrieval efficiency and boost the agent’s performance and success rates in constructing complex structures\cite{wang2023jarvis}. To address data scarcity, text-based prompt generators could be utilized to produce creative and detailed building instructions, enriching the memory pool with varied examples. Furthermore, integrating image-to-3D modeling tools, such as Tripo AI, could allow the memory to store voxel-based blueprints, enabling the agent to reason spatially and replicate intricate structures from visual inputs.

Another challenge stems from the limited range of primitive actions implemented in the current framework. This restriction prevents the agent from replicating more complex player behaviors, such as activating blocks, pouring liquids, or handling intricate mechanisms. Enhancing the primitive action set would enable the agent to undertake a broader range of construction tasks, such as the creation of functional semi-automated farms. Integrating downstream text-to-behavior controllers and policy execution frameworks, such as STEVE-1 or Voyager‌\cite{lifshitz2023steve,wang2024voyager}, could further optimize task execution and expand the agent’s behavioral repertoire.

Lastly, our findings indicate that the LLM-driven agent struggles to reason directly from visual references to produce accurate descriptive instructions and map these into precise blueprint layouts. This limitation highlights an opportunity for improvement through fine-tuning LLMs with training datasets that pair visual references with generated blueprints‌\cite{wei2022finetuned}. Such advancements could significantly enhance the agent’s ability to perform spatial reasoning and visualization tasks, capabilities not innately present in current LLMs but vital for more complex applications.

\section{Conclusion}
In this paper, we propose APT, an LLM-driven agent framework capable of constructing complex structures by leveraging the inherent reasoning abilities of large language models (LLMs) in both textual and spatial contexts. This framework is the first of its kind to integrate memory and reflection components into structure-building agents. While previous work has primarily focused on skill acquisition and task progression within the technology tree, the domain of structure building—a behavior highly characteristic of real player interactions in Minecraft—remains relatively unexplored. Our APT agent is capable of following extensive descriptive instructions to construct structures with detailed internal designs and arranged item layouts. Our findings demonstrate that the memory module significantly enhances the performance of few-shot and zero-shot learning agents. Conversely, the reflection module has shown limited impact. However, further advancements in cognitive framework for reflection\cite{wang2023describe} and the incorporation of enhanced computer vision capabilities could improve its effectiveness\cite{shinn2023reflexionlanguageagentsverbal,yao2023react}.

Additionally, we created a benchmarking dataset to evaluate agents’ abilities across creativity and spatial reasoning tasks. This benchmark provides a valuable tool for researchers to test diverse skill sets within this domain. Lastly, the agent's unexpected use of scaffolding—a technique widely employed in both real-world construction and survival-mode Minecraft gameplay—raises questions about the boundaries of emergent reasoning in LLMs and their ability to generalize human-like problem-solving techniques. Future research may also consider scaffolding modules for subroutine planning,  particularly in advancing the construction of more complex and sophisticated structures.

\bibliography{aaai25}

\section{Appendix}
This section provides the implementation details of our Chain-of-Thought (CoT) LLM framework and detailed examples of the prompts.
\subsection{Overview}
To fully utilize the potential of GPT-based LLMs for imagining and planning blueprints with spatial reasoning and precise item placements, we employed GPT-4o, GPT-4o Mini, GPT-3.5 Turbo, and GPT-O1. Through prompt engineering and a Chain-of-Thought (CoT) framework, we aimed to enhance the agent's ability to generate structured plans and execute them effectively.
\subsection{Layout Synopsis Generation Module}
\begin{tcolorbox}[
    colback=gray!5!white,     
    colframe=black!75!white, 
    title=Layout Synopsis Prompt, 
    sharp corners,           
    fonttitle=\bfseries,breakable    
]
\textbf{Please translate the structure in the provided text with the following Minecraft details:}

\begin{enumerate}
    \item \textbf{Components and Positioning:} List all individual elements (e.g., blocks, materials, windows, doors, etc.) used in the structure and describe the position of each component relative to the entire structure.
    \item \textbf{Dimensional Layout:} Provide the overall dimensions of the structure (length, width, height).
    \item \textbf{Description:} Summarize the purpose and design of the structure (e.g., a house, tower, etc.), and outline the most logical construction sequence, taking into account how building certain parts first could obstruct access to other areas.
\end{enumerate}

Please ensure the description is clear, precise, and professional, making it easy to recreate the structure programmatically.

\textbf{Here is the provided text description:} \texttt{\{text\}}
\end{tcolorbox}

\subsection{Blueprint Generation Module}%
Following the Layout Synopsis, the Blueprint Generation Module generates executable Python code to construct the layout within the Minecraft environment. This stage integrates a highly structured prompt to guide the LLM in producing code efficiently and dynamically.
\begin{tcolorbox}[
    colback=gray!5!white,     
    colframe=black!75!white, 
    title=Blueprint System Prompt, 
    sharp corners,           
    fonttitle=\bfseries,breakable
]
You are an expert in both Minecraft and Python coding. Your task is to
generate Python code that creates layouts for Minecraft structures as a
list of tuples.

The structure layout should be represented in the following way:

\begin{itemize}
    \item Each tuple contains:
    \begin{enumerate}
        \item The block type (e.g., \texttt{'oak\_planks'}, \texttt{'glass\_pane'}, \texttt{'oak\_door'}).
        \item The exact 3D position of the block, represented by a \texttt{vec3} object with \texttt{x}, \texttt{y}, and \texttt{z} coordinates.
    \end{enumerate}
\end{itemize}

The layout should follow this format:
\begin{lstlisting}[language=Python,numbers=none]
[
    ('block_type', start_pos.offset(x, y, z)),
    ('block_type', start_pos.offset(x, y, z)),
    ('block_type', start_pos.offset(x, y, z))
]
\end{lstlisting}

\textbf{Important Notes:}
\begin{itemize}
    \item You do not need to manually define every block's coordinates.
    Instead, provide efficient and reusable code that generates the layout
    dynamically, based on the starting position.
    \item The variable storing the layout must be named \texttt{layout}.
    \item Do not append \texttt{'air'} to the layout.
\end{itemize}

\# Always start the code with this:
\begin{lstlisting}[language=Python,numbers=none]
start_pos = self.bot.entity.position.floor()
\end{lstlisting}

\# Layout generation code that returns the layout

\# ...

\# The output always ends with the layout being passed into the following method:
\begin{lstlisting}[language=Python,numbers=none]
self.actions.buildStructure(layout, mode='creative')
\end{lstlisting}

\end{tcolorbox}
For the user’s input, we provide the retrieved plan from our RAG memory pool with the highest similarity score, along with the layout synopsis from the first module.

\begin{tcolorbox}[colback=gray!5!white,     
    colframe=black!75!white, 
    title=Blueprint User Prompt, 
    sharp corners,           
    fonttitle=\bfseries,breakable]
    Here is a possibly relevant plan from past experiences that were effective under their contexts (could be None). If the task described isn't relevant, don't reference it. If it is highly relevant, then reference it:
    {retrievedPlans}
    
    Please provide the code for this structure: {structure}.
    
    \begin{lstlisting}[language=Python,numbers=none]
    1. Do not import mineflayer or vec3
    2. Do not miss any components
    Ensure the generated code is properly indented and formatted as a complete Python script.
    \end{lstlisting}
    
    Only include the code and output the code into a compact JSON format on a single line without whitespace. The key is 'code' and the value is the actual code.
\end{tcolorbox}
\subsection{Reflections Module}
Once the agent completes the visual inspection of the constructed structure, the captured screenshots are seamlessly integrated into the self-reflection chain. This process combines the screenshots with the original structure description and the previously failed code. The prompt in the self-reflection chain generates not only the corrected code but also a comprehensive analysis of what was incorrect or suboptimal in the initial code. This analysis includes detailed insights into the generated structure, identification of potential errors, and recommendations for improvement, ensuring iterative enhancement and accuracy in subsequent builds.

\begin{tcolorbox}[colback=gray!5!white, 
    colframe=black!75!white,           
    title=Self Reflection Prompt,            
    sharp corners,                     
    fonttitle=\bfseries,breakable]               

You are an expert in both Minecraft structure generation and Python coding. Below is the current imperfect blueprint code used to generate a Minecraft structure, along with a description of the intended structure and an image of what was generated by the current code.

\textbf{Structure Description:}
{structure}

\textbf{Image:}
{Image}

\textbf{Current Blueprint Code:}
{blueprint}

\textbf{Task:} Generate the following features:
\begin{enumerate}
    \item \textbf{Reflection:} Analyze why the current blueprint code does not successfully generate the structure as described. Compare the structure description with both the image and the code itself. Issues may arise either from discrepancies in the visual appearance of the generated structure compared to the description, or from errors in the code's syntax that prevent it from running. Keep this concise.
    \item \textbf{Code:} Provide an improved, optimized version of the blueprint code that accurately aligns with the structure description and resolves any issues in the current code.
\end{enumerate}

\texttt{\# Always start the code with this:}
\begin{lstlisting}[language=Python]
start_pos = self.bot.entity.position.floor()
# layout generation code that returns the layout
# ...
# The output always ends with the layout being passed into the following method
self.actions.buildStructure(layout, mode='creative')
\end{lstlisting}

Ensure the generated code is properly indented and formatted as a complete Python script. Include the code and reasoning into a compact JSON format on a single line without whitespace where we have two keys "reflection" and "code" and the value is the corresponding output.

\end{tcolorbox}

%

\end{document}